\documentclass[runningheads]{llncs}
\usepackage{graphicx}

\usepackage{tikz}
\usepackage{comment}
\usepackage{amsmath,amssymb} 
\usepackage{color}

\usepackage{multirow}
\usepackage{microtype}
\usepackage[caption=false]{subfig}
\usepackage{booktabs}

\usetikzlibrary{arrows,intersections}
\tikzset{font=\scriptsize}
\usepackage{pgfplots}
\pgfplotsset{compat=1.11}
\usepgfplotslibrary{fillbetween}
\usetikzlibrary{backgrounds}

\usepackage[accsupp]{axessibility}  

\usepackage[pagebackref=true,breaklinks=true,colorlinks,bookmarks=false]{hyperref}

\begin{document}
\pagestyle{headings}
\mainmatter
\def\ECCVSubNumber{xxxx}  

\title{Exploiting Unlabeled Data with Vision and Language Models for Object Detection} 

\titlerunning{Exploiting Unlabeled Data with V\&L Models}
%
\author{Shiyu Zhao\inst{1,}\thanks{Equal contribution.} \and
Zhixing Zhang\inst{1,\star} \and
Samuel Schulter\inst{2} \and 
Long Zhao\inst{3} \and \\
Vijay Kumar B.G\inst{2}\index{B.G, Vijay Kumar} \and 
Anastasis Stathopoulos\inst{1} \and 
Manmohan Chandraker\inst{2,4} \and 
Dimitris Metaxas\inst{1}
}
\authorrunning{S. Zhao \emph{et al.}}
%
\institute{\textsuperscript{1}Rutgers University, 
\textsuperscript{2}NEC Labs America, 
\textsuperscript{3}Google Research, 
\textsuperscript{4}UC San Diego
}

\maketitle

\begin{abstract}
    Building robust and generic object detection frameworks requires scaling to larger label spaces and bigger training datasets. However, it is prohibitively costly to acquire annotations for thousands of categories at a large scale.  We propose a novel method that leverages the rich semantics available in recent vision and language models to localize and classify objects in unlabeled images, effectively generating pseudo labels for object detection.  Starting with a generic and class-agnostic region proposal mechanism, we use vision and language models to categorize each region of an image into any object category that is required for downstream tasks.  We demonstrate the value of the generated pseudo labels in two specific tasks, open-vocabulary detection, where a model needs to generalize to unseen object categories, and semi-supervised object detection, where additional unlabeled images can be used to improve the model.  Our empirical evaluation shows the effectiveness of the pseudo labels in both tasks, where we outperform competitive baselines and achieve a novel state-of-the-art for open-vocabulary object detection. Our code is available at \url{https://github.com/xiaofeng94/VL-PLM}.
\end{abstract}

\section{Introduction}

Recent advances in object detection build on large-scale datasets~\cite{gupta2019lvis,OpenImages,Objects365}, which provide rich and accurate human-annotated bounding boxes for many object categories.
However, the annotation cost of such datasets is significant.  Moreover, the long-tailed distribution of natural object categories makes it even harder to collect sufficient annotations for all categories. 
Semi-supervised object detection (SSOD)~\cite{sohn2020detection,zhou_cvpr_21} and open-vocabulary object detection (OVD)~\cite{bansal2018zero,gu_iclr_22,zareian_cvpr_21} are two tasks to lower annotations costs by leveraging different forms of unlabeled data.  In SSOD, a small fraction of fully-annotated training images is given along with a large corpus of unlabeled images.  In OVD, a fraction of the desired object categories is annotated (the base categories) in all training images and the task is to also detect a set of novel (or unknown) categories at test time.  These object categories can be present in the training images, but are not annotated with ground truth bounding boxes.  A common and successful approach for leveraging unlabeled data is by generating pseudo labels.  However, all prior works on SSOD only leveraged the small set of labeled data for generating pseudo labels, while most prior work on OVD does not leverage pseudo labels at all.

\begin{figure}[t]\centering
    \includegraphics[width=.95\linewidth]{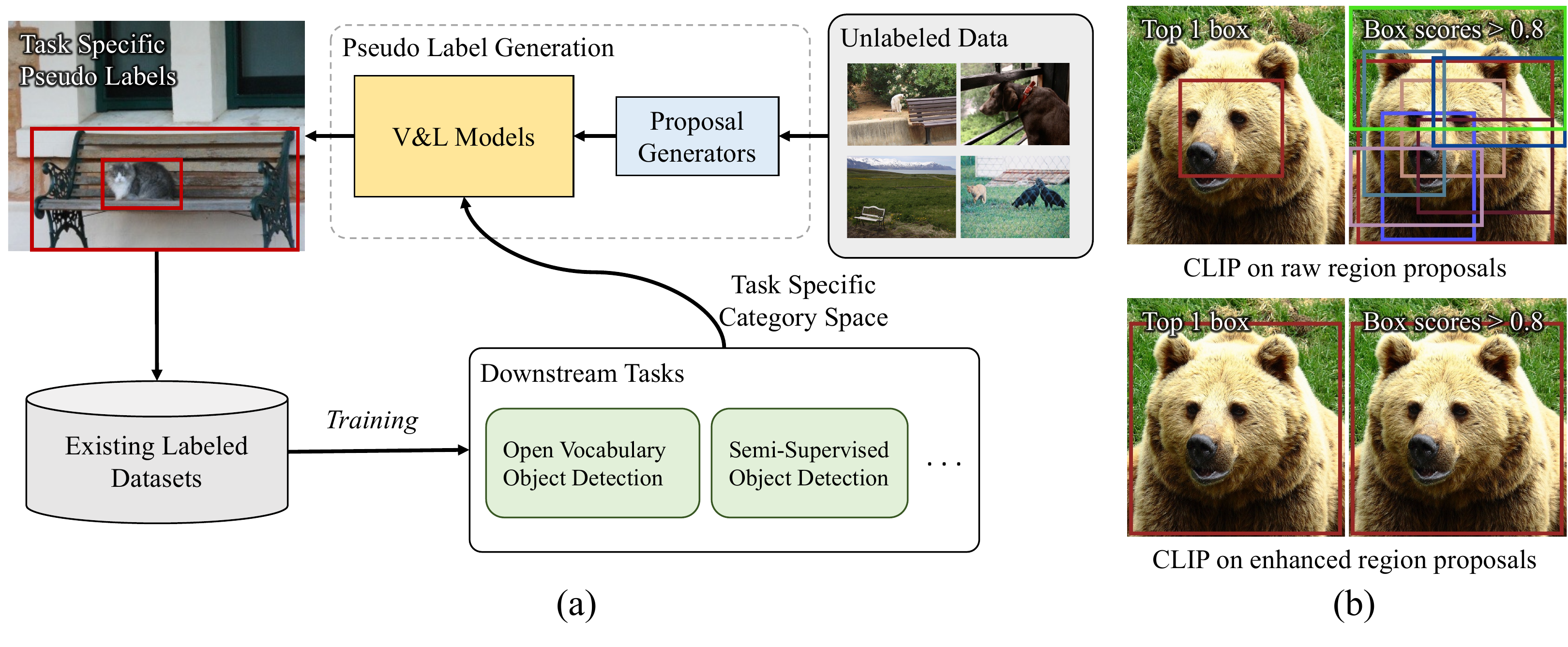}
    \caption{\textbf{(a)} Overview of leveraging the semantic knowledge contained in vision and language models for mining unlabeled data to improve object detection systems for open-vocabulary and semi-supervised tasks.
    \textbf{(b)} Illustration of the weak localization ability when applying CLIP~\cite{radford_arxiv_2021} on raw object proposals (top), compared with our improvements (bottom).  The left images show the pseudo label with the highest score.  The right images show all pseudo labels with scores greater than 0.8.  The proposed scoring gives much cleaner pseudo labels.
    }
    \label{fig:teaser}
\end{figure}

In this work, we propose a simple but effective way to mine unlabeled images using recently proposed vision and language (V\&L) models to generate pseudo labels for both known and unknown categories, which suits both tasks, SSOD and OVD.
V\&L models~\cite{jia_icml_21,ALBEF,radford_arxiv_2021} can be trained from (noisy) image caption pairs, which can be obtained at a large scale without human annotation efforts by crawling websites for images and their alt-texts.  Despite the noisy annotations, these models demonstrate excellent performance on various semantic tasks like zero-shot classification or image-text retrieval.  The large amount of diverse images, combined with the free-form text, provides a powerful source of information to train robust and generic models.
These properties make vision and language models an ideal candidate to improve existing object detection pipelines that leverage unlabeled data, like OVD or SSOD, see Fig.~\ref{fig:teaser}(a).

Specifically, our approach leverages the recently proposed vision and language model CLIP~\cite{radford_arxiv_2021} to generate pseudo labels for object detection.
We first predict region proposals with a two-stage class-agnostic proposal generator which was trained with limited ground truth (using only known base categories in OVD and only labeled images in SSOD), but generalizes to unseen categories.  For each region proposal, we then obtain a probability distribution over the desired object categories (depending on the task) with the pre-trained V\&L model CLIP~\cite{radford_arxiv_2021}.
However, as shown in Fig.~\ref{fig:teaser}(b), a major challenge of V\&L models is the rather low object localization quality, also observed in~\cite{zhong2021regionclip}. 
To improve localization, we propose two strategies where the two-stage proposal generator helps the V\&L model: (1) Fusing CLIP scores and objectness scores of the two-stage proposal generator, and (2) removing redundant proposals by repeated application of the localization head (2nd stage) in the proposal generator.
Finally, the generated pseudo labels are combined with the original ground truth to train the final detector.  We name our method as \textbf{V}\&\textbf{L}-guided \textbf{P}seudo-\textbf{L}abel \textbf{M}ining (VL-PLM).

Extensive experiments demonstrate that VL-PLM successfully exploits the unlabeled data for open-vocabulary detection and outperforms the state-of-the-art ViLD~\cite{gu_iclr_22} on novel categories by +6.8 AP on the COCO dataset~\cite{COCO}.
Moreover, VL-PLM improves the performance on known categories in SSOD and beats the popular baseline STAC~\cite{sohn2020detection} by a clear margin, by only replacing its pseudo labels with ours. 
Besides, we also conduct various ablation studies on the properties of the generated pseudo labels and analyze the design choices of our proposed method.
We also believe that VL-PLM can be further improved with better V\&L models like ALIGN~\cite{jia_icml_21} or ALBEF~\cite{ALBEF}.

The contributions of our work are as follows:
\textbf{(1)} We leverage V\&L models for improving object detection frameworks by generating pseudo labels on unlabeled data.
\textbf{(2)} A simple but effective strategy to improve the localization quality of pseudo labels scored with the V\&L model CLIP~\cite{radford_arxiv_2021}.
\textbf{(3)} State-of-the-art results for novel categories on the COCO open-vocabulary detection setting.
\textbf{(4)} We showcase the benefits of VL-PLM in a semi-supervised object detection setting.

\section{Related Work}
\label{sec:related_work}

The goal of our work is to improve object detection systems by leveraging unlabeled data via vision and language models that carry rich semantic information.

\vspace{1mm}
\noindent \textbf{Vision \& language (VL) models:}
Combining natural language and images has enabled many valuable applications in recent years, like image captioning~\cite{agrawal_iccv_19,chen2015microsoft,fang_cvpr_15,karpathy_cvpr_15}, visual question answering~\cite{agrawal_iccv_15,fukui_emnlp_16,hudson_neurips_19,li_eccv_20,peng_cvpr_2019,zhang_cvpr_21}, referring expression comprehension~\cite{Chen_ECCV_20_UNITER,kamath_iccv_21,kazemzadeh_emnlp_14,Lu_neurips_19_ViLBERT,mao_cvpr_16,yu_cvpr_18,yu_eccv_16}, image-text retrieval~\cite{ALBEF,radford_arxiv_2021,wang_cvpr_16} or language-driven embodied AI~\cite{anderson_cvpr_18,das_cvpr_18}.
While early works proposed task-specific models, generic representation learning from vision and language inputs has gained more attention~\cite{Chen_ECCV_20_UNITER,hu_iccv_21_UniT,liu2019roberta,Lu_neurips_19_ViLBERT,Sun_iccv_19_VideoBERT}.
Most recent works like CLIP~\cite{radford_arxiv_2021} or ALIGN~\cite{jia_icml_21} also propose generic vision and language representation learning approaches, but have significantly increased the scale of training data, which led to impressive results in tasks like zero-shot image classification or image-text retrieval. The training data consist of image and text pairs, typically crawled from the web at a very large scale (400M for \cite{radford_arxiv_2021} and 1.2B for \cite{jia_icml_21}), but without human annotation effort.
In our work, we leverage such pre-trained models to mine unlabeled data and to generate pseudo labels in the form of bounding boxes, suitable for object detection.  One challenge with using such V\&L models~\cite{jia_icml_21,radford_arxiv_2021} is their limited capability in localizing objects (recall Fig.~\ref{fig:teaser}(b)), likely due to the lack of region-word alignment in the image-text pairs of their training data.  In Sec.~\ref{sec:method_pl_with_vl}, we show how to improve localization quality with our proposal generator.

\vspace{1mm}
\noindent \textbf{Vision \& language models for dense prediction tasks:} 
The success of CLIP~\cite{radford_arxiv_2021} (and others~\cite{jia_icml_21,ALBEF}) has motivated the extension of zero-shot classification capabilities to dense image prediction tasks like object detection~\cite{gu_iclr_22,huynh2021openvocabulary,shi2022proposalclip,zareian_cvpr_21} or semantic segmentation~\cite{li_iclr_22,rao2021denseclip,xu2021simple,zhou2021denseclip}.
These works try to map features of individual objects (detection) or pixels (segmentation) into the joint vision-language embedding space provided by models like CLIP.
For example, ViLD~\cite{gu_iclr_22} trains an object detector in the open-vocabulary regime by predicting the text embedding (from the CLIP text-encoder) of the category name for each image region. 
LSeg~\cite{li_iclr_22} follows a similar approach, but is applied to zero-shot semantic segmentation.  Both works leverage task-specific insights and do not generate explicit pseudo labels.  In contrast, our proposed VL-PLM is more generic by generating pseudo labels, thus enabling also other tasks like semi-supervised object detection~\cite{sohn2020detection}.
Similar to our work, both Gao \emph{et al.}~\cite{gao2021open} and Zhong \emph{et al.}~\cite{zhong2021regionclip} generate explicit pseudo labels in the form of bounding boxes.
In \cite{gao2021open}, the attention maps of a pretrained V\&L model~\cite{ALBEF} between words of a given caption and image regions are used together with object proposals to generate pseudo labels.  
In contrast, our approach does not require image captions as input and we use only unlabeled images, while still outperforming~\cite{gao2021open} in an open-vocabulary setting on COCO.  
RegionCLIP \cite{zhong2021regionclip} assigns semantics to region proposals via a pre-trained V\&L model, effectively creating pseudo labels in the form of bounding boxes. 
While our approach uses such pseudo labels directly for training object detectors, \cite{zhong2021regionclip} uses them for fine-tuning the original V\&L model, which then builds the basis for downstream tasks like open-vocabulary detection.  
We believe this contribution is orthogonal to ours as it effectively builds a better starting point of the V\&L model, and can be incorporated into our framework as well. Interestingly, even without the refined V\&L model, we show improved accuracy with pseudo labels specifically for novel categories as shown in Sec.~\ref{sec:exp_ovd}.

The main focus of all the aforementioned works is to enable the dynamic expansion of the label space and to recognize novel categories.  While our work also demonstrates state-of-the-art results in this open-vocabulary setting, where we mine unlabeled data for novel categories, we want to stress that our pseudo labels are applicable more generally.  In particular, we also use a V\&L model to mine unlabeled images for known categories in a semi-supervised object detection setting.
Furthermore, by building on the general concept of pseudo labels, our approach may be extended to other dense prediction tasks like semantic segmentation in future works as well.

\vspace{1mm}
\noindent \textbf{Object detection from incomplete annotations:}
Pseudo labels are proven useful in many recent object detection methods trained with various forms of weak annotations: semi-supervised detection~\cite{sohn2020detection,zhou_cvpr_21}, unsupervised object discovery~\cite{LOST}, open-vocabulary detection~\cite{gao2021open,zhong2021regionclip}, weakly-supervised detection~\cite{dong2021boosting,zhong2020boosting}, unsupervised domain adaptation~\cite{inoue_cvpr_18,yu_wacv_22} or multi-dataset detection~\cite{zhao_eccv_20}.  In all cases, an initial model trained from base information is applied on the training data to obtain the missing information.
Our main proposal is to leverage V\&L models to improve these pseudo labels and have one unified way of improving the accuracy in multiple settings, see Sec.~\ref{sec:method_pl_for_specific_tasks}.  In this work, we focus on two important forms of weak supervision: zero-shot/open-vocabulary detection (OVD) and semi-supervised object detection (SSOD).
In zero-shot detection~\cite{bansal2018zero} a model is trained from a set of base categories.  Without ever seeing any instance of a novel category during training, the model is asked to predict novel categories, typically via association in a different embedding space, like attribute or text embeddings.  Recent works~\cite{gu_iclr_22,rahman2020improved,zareian_cvpr_21} relax the setting to include novel categories in the training data, but without bounding box annotations, which also enables V\&L models to be used (via additional images that come with caption data).  ViLD~\cite{gu_iclr_22}, as described above, uses CLIP~\cite{radford_arxiv_2021} with model distillation losses to make predictions in the joint vision-text embedding space.  In contrast, we demonstrate that explicitly creating pseudo labels for novel categories via mining the training data can significantly improve the accuracy, see Sec.~\ref{sec:exp_ovd}.
The second task we focus on is semi-supervised object detection (SSOD), where a small set of images with bounding box annotations and a large set of unlabeled images are given.  In contrast to OVD, the label space does not change from train to test time.
A popular and recent baseline that builds on pseudo labels is STAC~\cite{sohn2020detection}.  This approach employs a consistency loss between predictions on a strongly augmented image and pseudo labels computed on the original image.  We demonstrate the benefit of leveraging V\&L models to improve the pseudo label quality in such a framework.  Other works on SSOD, like~\cite{xu_iccv2021_softteacher,zhou_cvpr_21} propose several orthogonal improvements which can be incorporated into our framework as well.  In this work, however, we focus purely on the impact of the pseudo labels.
Finally, note that our concepts may also be applicable to other tasks beyond open-vocabulary and semi-supervised object detection, but we leave this for future work.

\section{Method}
\label{sec:method}

The goal of our work is to mine unlabeled images with vision \& language (V\&L) models to generate semantically rich pseudo labels (PLs) in the form of bounding boxes so that object detectors can better leverage unlabeled data.  
We start with a generic training strategy for object detectors with the unlabeled data in Sec.~\ref{sec:method_train_from_unlabeled}. Then, Sec.~\ref{sec:method_pl_with_vl} describes the proposed VL-PLM for pseudo label generation. 
Finally, Sec.~\ref{sec:method_pl_for_specific_tasks} presents specific object detection tasks with our PLs.

\subsection{Training object detectors with unlabeled data}
\label{sec:method_train_from_unlabeled}

Unlabeled data comes in many different forms for object detectors.  In semi-supervised object detection, we have a set of fully-labeled images $\mathcal{I}_L$ with annotations for the full label space $\mathcal{S}$, as well as unlabeled images $\mathcal{I}_U$, with  $\mathcal{I}_L \cap \mathcal{I}_U = \varnothing$.  In open-vocabulary detection, we have partly-labeled images with annotations for the set of base categories $\mathcal{S}_B$, but without annotations for the unknown/novel categories $\mathcal{S}_N$.  Note that partly-labeled images are therefore contained in both $\mathcal{I}_L$ and $\mathcal{I}_U$, i.e., $\mathcal{I}_L = \mathcal{I}_U$.

A popular and successful approach to learn from unlabeled data is via pseudo labels.  Recent semi-supervised object detection methods follow this approach by first training a teacher model on the limited ground truth data, then generating pseudo labels for the unlabeled data, and finally training a student model.  In the following, we describe a general training strategy for object detection to handle different forms of unlabeled data.

We define a generic loss function for an object detector with parameters $\theta$ over both labeled and unlabeled images as
\begin{align}
    \mathcal{L}(\theta, \mathcal{I}) = \frac{1}{N_{\mathcal{I}}} \sum_{i=1}^{N_{\mathcal{I}}} [I_i \in \mathcal{I}_L] \; l_s (\theta, I_i) + \alpha [I_i \in \mathcal{I}_U] \; l_u (\theta, I_i) \;,
\end{align}
where $\alpha$ is a hyperparameter to balance supervised $l_s$ and unsupervised $l_u$ losses and $[\cdot]$ is the indicator function returning either 0 or 1 depending on the condition.  Note again that $I_i$ can be contained in both $\mathcal{I}_L$ and $\mathcal{I}_U$.

Object detection ultimately is a set prediction problem and to define a loss function, the set of predictions (class probabilities and bounding box estimates) need to be matched with the set of ground truth boxes.  Different options exist to find a matching~\cite{carion_eccv_20_detr,he_iccv_17_maskrcnn} but it is mainly defined by the similarity (IoU) between predicted and ground truth boxes.  We define the matching for prediction $i$ as $\sigma(i)$, which returns a ground truth index $j$ if successfully matched or \texttt{nil} otherwise.
The supervised loss $l_s$ contains a standard cross-entropy loss for the classification $l_{cls}$ and an $\ell_1$ loss for the box regression $l_{reg}$.  Given $I \in \mathcal{I}$, we define $l_s$ as,
\begin{align}
    l_s (\theta, I) = \frac{1}{N^*} \sum_{i} l_{cls}\left(C_i^\theta (I), c^*_{\sigma(i)}\right) + [\sigma(i) \neq \texttt{nil}] \; l_{reg}\left(T_i^\theta (I), \mathbf{t}^*_{\sigma(i)}\right) \;,
\end{align}
where $N^*$ is the number of predicted bounding boxes.  $C_i^\theta(\cdot)$ and $T_i^\theta(\cdot)$ are the predicted class distributions and bounding boxes of the object detector. The corresponding (matched) ground truth is defined as $c^*_{\sigma(i)}$ and $\mathbf{t}^*_{\sigma(i)}$, respectively.

The unsupervised loss $l_u$ is similarly defined, but uses pseudo labels with high confidence as supervision signals:
\begin{align}
\begin{split}
    l_u (\theta, I) = \frac{1}{N^u} \sum_{i} \; [\max(\mathbf{p}^u_{\sigma(i)}) \geq \tau] \; \cdot & \left( l_{cls}\left(C_i^\theta (I), \hat{c}^u_{\sigma(i)}\right) + \right. \\
    &\left.\; [\sigma(i) \neq \texttt{nil}] \; l_{reg}\left(T_i^\theta (I), \mathbf{t}^u_{\sigma(i)}\right) \right) \;.
\end{split}
\end{align}
Here, $\mathbf{p}^u_{\sigma(i)}$ defines the probability distribution over the label space of the pseudo label matched with prediction $i$ and $N^u$ is the number of adopted pseudo labels, i.e., $N^u = \sum_i [\max(\mathbf{p}^u_{\sigma(i)}) \geq \tau]$. 
Pseudo labels for the classification and the box regression losses are $\hat{c}^u_{\sigma(i)} = \arg\max (\mathbf{p}^u_{\sigma(i)})$ and $\mathbf{t}^u_{\sigma(i)}$, respectively.

The key to successful training of object detectors from unlabeled data are accurate pseudo labels.  In the next section, we will present our approach, VL-PLM, to leverage V\&L models as external models to exploit unlabeled data for generating pseudo labels.

\begin{figure}[t]\centering
    \includegraphics[width=1\linewidth]{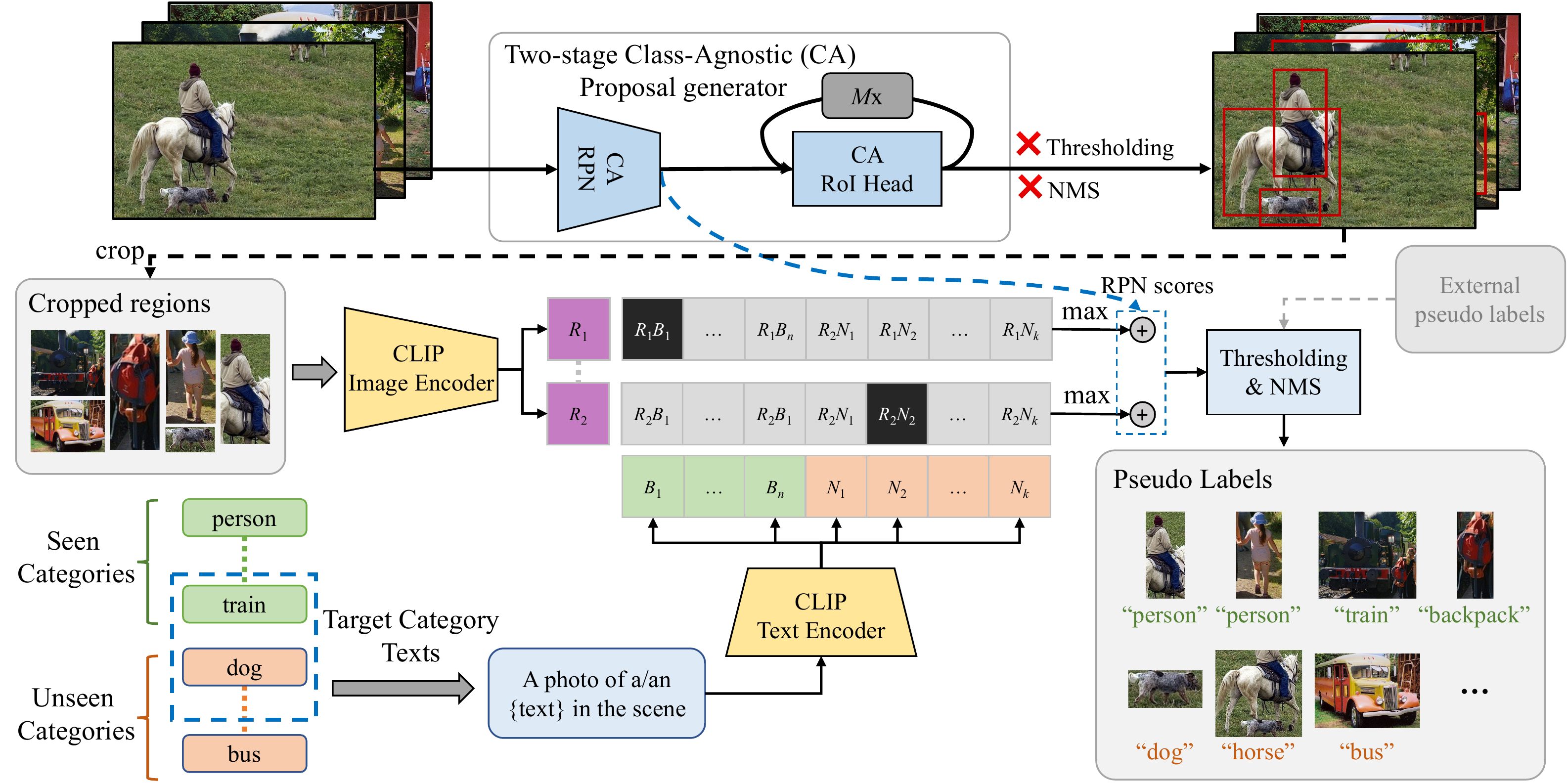}
    \caption{
    Overview of the proposed VL-PLM to mine unlabeled images with vision \& language models to generate pseudo labels for object detection.  The top part illustrates our class-agnostic proposal generator, which improves the pseudo label localization by using the class-agnostic proposal score and the repeated application of the RoI head.  The bottom part illustrates the scoring of cropped regions with the V\&L model based on the target category names.  The chosen category names can be adjusted for the desired downstream task.  After thresholding and NMS, we get the final pseudo labels.  For some tasks like SSOD, we will merge external pseudo labels for a teacher model with ours before thresholding and NMS.
    }
    \label{fig:overview_pipeline}
\end{figure}

\begin{figure}[t]\centering
    \includegraphics[width=1\linewidth]{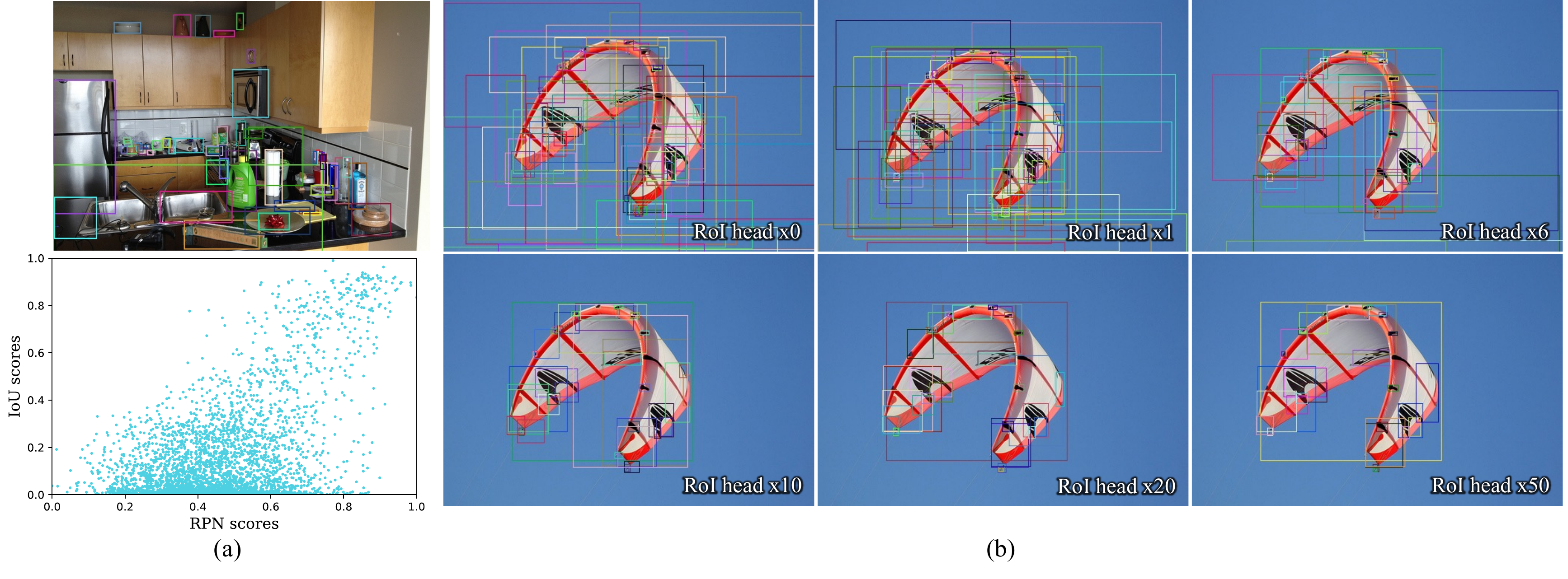}
    \caption{\textbf{(a)} RPN scores indicate localization quality. Top: Top 50 boxes from RPN in an image which correctly locates nearly all objects. Bottom: A positive correlation between RPN and IoU scores for RPN boxes of 50 randomly sampled COCO images. The correlation coefficient is 0.51.
    \textbf{(b)} Box refinement by repeating RoI head. ``$\times$N'' indicates how many times we repeat the RoI head.}
    \label{fig:method_localization}
\end{figure}

\subsection{VL-PLM: Pseudo labels from vision \& language models}
\label{sec:method_pl_with_vl}
V\&L models are trained on large scale datasets with image-text pairs that cover a diverse set of image domains and rich semantics in natural text.  Moreover, the image-text pairs can be obtained without costly human annotation by using web-crawled data (images and corresponding alt-texts)~\cite{radford_arxiv_2021,jia_icml_21}.   Thus, V\&L models are ideal sources of external knowledge to generate pseudo labels for arbitrary categories, which can be used for downstream tasks like open-vocabulary or semi-supervised object detection. 

\vspace{1mm}
\noindent \textbf{Overview:}
Fig.~\ref{fig:overview_pipeline} illustrates the overall pipeline of our pseudo label generation with the recent V\&L model CLIP~\cite{radford_arxiv_2021}.  We first feed an unlabeled image into our two-stage class-agnostic detector (described in the next section below) to obtain region proposals.  We then crop image patches based on those regions and feed them into the CLIP image-encoder to obtain an embedding in the CLIP vision-and-language space.  Using the corresponding CLIP text-encoder and template text prompts, we generate embeddings for category names that are desired for the specific task.
For each region, we compute the similarities between the region embedding and the text embeddings via a dot product and use softmax to obtain a distribution over the categories. 
We then generate the final pseudo labels using scores from both class-agnostic detector and V\&L model, which we describe in detail below.

There are two key challenges in our framework: (1) Generating robust proposals for novel categories, required by open-vocabulary detection, and (2) overcoming the poor localization quality of the raw CLIP model, see Fig.~\ref{fig:teaser}(b).  We introduce simple but effective solutions to address the two challenges in the following.

\vspace{1mm}  
\noindent \textbf{Generating robust and class-agnostic region proposals:}
To benefit tasks like open vocabulary detection with the unlabeled data, the proposal generator should be able to locate not only objects of categories seen during training but also of objects of novel categories.
While unsupervised candidates like selective search~\cite{Uijlings13_selectivesearch} exist, these are often time-consuming and generate many noisy boxes.
As suggested in prior studies \cite{gu_iclr_22,zareian_cvpr_21}, the region proposal network (RPN) of a two-stage detector generalizes well for novel categories. 
Moreover, we find that the RoI head is able to improve the localization of region proposals, which is elaborated in the next section.
Thus, we train a standard two-stage detector, e.g., Faster-RCNN~\cite{ren_neurips15_fasterrcnn}, as our proposal generator using available ground truth, which are annotations of base categories for open vocabulary detection and annotations from the small fraction of annotated images in semi-supervised detection. 
To further improve the generalization ability, we ignore the category information of the training set and train a class-agnostic proposal generator.
Please refer to Sec.~\ref{subsec:abstudy} and the supplement for a detailed analysis of the proposal generator.

\vspace{1mm}
\noindent \textbf{Generating pseudo labels with a V\&L model:}
Directly applying CLIP~\cite{radford_arxiv_2021} on cropped region proposals yields low localization quality, as was observed in Fig.~\ref{fig:teaser}(b) and also in~\cite{zhong2021regionclip}.
Here, we demonstrate how to improve the localization ability with our two-stage class-agnostic proposal generator in two ways.
Firstly, we find that the RPN score is a good indicator for localization quality of region proposals.  Fig.~\ref{fig:method_localization}(a) illustrates a positive correlation between RPN and IoU scores.  We leverage this observation and average the RPN score with those of the CLIP predictions.
Secondly, we remove thresholding and NMS of the proposal generator and feed proposal boxes into the RoI head multiple times, similar to~\cite{cai18cascadercnn}.
We observe that it pushes redundant boxes closer to each other by repeating the RoI head, which can be seen in Fig.~\ref{fig:method_localization}(b).
In this way, we encounter better located bounding boxes and provide better pseudo labels.  Please refer to Sec.~\ref{subsec:abstudy} for a corresponding empirical analysis.

To further improve the quality of our pseudo labels, we adopt the multi-scale region embedding from CLIP as described in~\cite{gu_iclr_22}.  Moreover, as suggested in~\cite{sohn2020detection}, we employ a high threshold to pick pseudo labels with high confidence.
The confidence score of the pseudo label for the region $R_i$ is formulated as $\overline{c}^{u}_i = [s^{u}_i \geq \tau] \cdot s^{u}_i$, with
\begin{align}
    s^{u}_i &= \frac{ S_{RPN} (R_i) + \max(\mathbf{p}^{u}_i ) }{2} \;,
\end{align}
where $S_{RPN}(\cdot)$ denotes the RPN score. The prediction probability distribution $\mathbf{p}^{u}_i$ is defined as
\begin{align}
    \mathbf{p}^{u}_i &= \text{softmax}\{ \phi( E_{\text{im}} (R_i) + E_{\text{im}} (R_i^{1.5\times})) \cdot E_{\text{txt}} (\text{Categories})^T \} .
\end{align}
Here, $R_i^{1.5\times}$ is a region cropped by $1.5\times$ the size of $R_i$. $E_{\text{im}}$ and $E_{\text{txt}}$ are the image and text encoders of CLIP, respectively, and $\phi(\mathbf{x}) = \mathbf{x}/||\mathbf{x}||$.
If $\overline{c}^{u}_i = 0$, we exclude $R_i$ from our pseudo labels.

\subsection{Using our pseudo labels for downstream tasks}
\label{sec:method_pl_for_specific_tasks}

Finally, we briefly describe how we use the pseudo labels that are generated from unlabeled data for two specific downstream tasks that we focus on in this work.

\vspace{1mm}
\noindent \textbf{Open-vocabulary detection:}
In this task, the detector has access to images with annotations for base categories and needs to generalize to novel categories.  
We leverage the data of the base categories to train a class-agnostic Mask R-CNN as our proposal generator and take the names of novel categories as the input texts of the CLIP text-encoder in aforementioned pseudo label generation process.  
Then, we train a standard Mask R-CNN with RestNet50-FPN~\cite{lin_cvpr_17_fpn} with both base ground truth and novel pseudo labels as described in Sec.~\ref{sec:method_train_from_unlabeled}.

\vspace{1mm}
\noindent \textbf{Semi-supervised object detection:}
In this task, relevant methods usually train a teacher model using ground truth from the limited set of labeled images, and then generate pseudo labels with the teacher on the unlabeled images.  We also generate those pseudo labels and merge them with pseudo labels from our VL-PLM.
Please refer to the supplementary document for details.
Thus, the student model is trained on available ground truth and pseudo labels from both our V\&L-based approach and the teacher model.

\section{Experiments}

We experimentally evaluate the proposed VL-PLM first on open-vocabulary detection in Sec.~\ref{sec:exp_ovd} and then on semi-supervised object detection in Sec.~\ref{sec:exp_ssod}.  In Sec.~\ref{subsec:abstudy} we ablate various design choices of VL-PLM.

\subsection{Open-vocabulary object detection}
\label{sec:exp_ovd}
In this task, we have a training set with annotations for known base categories $\mathcal{S}_B$. Our goal is to train a detector for novel categories $\mathcal{S}_N$. Usually, the labeled images $\mathcal{I}_L$ and the unlabeled images $\mathcal{I}_U$ are the same, i.e., $\mathcal{I}_L = \mathcal{I}_U$.

\vspace{1mm}
\noindent \textbf{Experimental setup:} 
Following prior studies \cite{bansal2018zero,gao2021open,gu_iclr_22,zareian_cvpr_21}, we base our evaluation on COCO 2017~\cite{COCO} in the zero-shot setting (COCO-ZS) where there are 48 known base categories and 17 unknown novel categories. 
Images from the training set are regarded as labeled for base classes and also as unlabeled for novel classes.
We take the widely adopted mean Average Precision at an IoU of 0.5 (AP$_{50}$) as the metric and mainly compare our method with ViLD~\cite{gu_iclr_22}, the state-of-the-art method for open vocabulary detection.
Thus, we follow ViLD and report AP$_{50}$ over novel categories, base categories and all categories as Novel AP, Base AP, and Overall AP, respectively.
Our supplemental material contains results for the LVIS~\cite{gupta2019lvis} dataset.

\begin{table}[tb]
\begin{center}
\caption{Evaluations for open vocabulary detection on the COCO 2017~\cite{COCO}. RegionCLIP* indicates a model without refinement using image-caption pairs.
}
\label{table:open_voc_results}
\resizebox{\textwidth}{!}{  
    \begin{tabular}{llcccl}
    \toprule
    Method & Training Source & Novel AP & { Base AP} & { Overall AP} &  \\
    \hline
    Bansal \emph{et al.}~\cite{bansal2018zero} &  & 0.31 &29.2 &  24.9 &  \\
    Zhu \emph{et al.}~\cite{zhu2020don} &  & 3.41 & 13.8 & 13.0 &  \\
    Rahman \emph{et al.}~\cite{rahman2020improved} & \multirow{-3}{*}{instance-level labels in $\mathcal{S}_B$} & 4.12 & 35.9 & 27.9 &  \\
    \hline
     & image-caption pairs in $\mathcal{S}_B \cup \mathcal{S}_N$ &  & { } & { } &  \\
    \multirow{-2}{*}{OVR-CNN~\cite{zareian_cvpr_21}} & instance-level labels in $\mathcal{S}_B$ & \multirow{-2}{*}{22.8} & \multirow{-2}{*}{46.0} & \multirow{-2}{*}{39.9} &  \\ \hline
     &  &  &  &  &  \\
     &  &  &  &  &  \\
    \multirow{-3}{*}{\begin{tabular}[l]{@{}l@{}}Gao \emph{et al.}~\cite{gao2021open} \\ RegionCLIP \cite{zhong2021regionclip}\end{tabular}} & \multirow{-3}{*}{\begin{tabular}[c]{@{}l@{}} raw image-text pairs via Internet\\ image-caption pairs in $\mathcal{S}_B \cup \mathcal{S}_N$\\ instance-level labels in $\mathcal{S}_B$\end{tabular}} & \multirow{-3}{*}{\begin{tabular}[c]{@{}l@{}}30.8\\31.4\end{tabular}} & \multirow{-3}{*}{\begin{tabular}[c]{@{}l@{}}46.1\\57.1\end{tabular}} & \multirow{-3}{*}{\begin{tabular}[c]{@{}l@{}}42.1\\50.4\end{tabular}} &  \\ \hline
    RegionCLIP* \cite{zhong2021regionclip} &  & 14.2 & 52.8 & 42.7 &  \\
    ViLD \cite{gu_iclr_22} &  & 27.6 & 59.5 & 51.3 &  \\
    VL-PLM (Ours) & \multirow{-3}{*}{\begin{tabular}[l]{@{}l@{}}raw image-text pairs via Internet\\ instance-level labels in $\mathcal{S}_B$\end{tabular}} & {\bf 34.4} & {\bf60.2} & {\bf 53.5} & \\
    \bottomrule
    \end{tabular}
}
\end{center}
\end{table}

\vspace{1mm}
\noindent \textbf{Implementation details:}
We set a NMS threshold of 0.3 for the RPN of the proposal generator. The confidence threshold for pseudo labels (PLs) is $\tau = 0.8$. Finally, we obtain an average of 4.09 PLs per image, which achieve a Novel AP of 20.9. We use the above hyperparameters for pseudo label generation in all experiments, unless otherwise specified.
The proposal generator and the final detector were implemented in Detectron2~\cite{wu2019detectron2} and trained on a server with NVIDIA A100 GPUs. The proposal generator was trained for 90,000 iterations with a batch size of 16. 
Similar to ViLD, the final detector is trained from scratch for 180,000 iterations with input size of $1024\times1024$, large-scale jitter augmentation \cite{ghiasi2021simple}, synchronized batch normalization of batch size 128, weight decay of 4e-5, and an initial learning rate of 0.32.

\vspace{1mm}
\noindent \textbf{Comparison to SOTA:} 
As shown in Table~\ref{table:open_voc_results}, the detector trained with VL-PLM significantly outperforms the prior state-of-the-art ViLD by nearly +7\% in Novel AP. 
Compared with \cite{zareian_cvpr_21} and \cite{gao2021open}, our method achieves much better performance not only on novel but also on base categories. This indicates training with our PLs has less impact on the predictions of base categories, where previous approaches suffered a huge performance drop.
Overall, we can see that using V\&L models to explicitly generate PLs for novel categories to train the model can give a clear performance boost.  Although this introduces an overhead compared to ViLD (and others), which can include novel categories dynamically into the label space, many practical applications easily tolerate this overhead in favor of significantly improved accuracy.
Such a setup is also similar to prior works that generate synthetic features of novel categories~\cite{zhu2020don}.
Moreover, our method has large potential for further improvement with better V\&L model.
\cite{gu_iclr_22} demonstrates a 60\% performance boost of ViLD when using ALIGN~\cite{jia_icml_21} as the V\&L model.
We expect similar improvements on VL-PLM if ALIGN is available.

\vspace{1mm}
\noindent \textbf{Generalizing to unseen datasets:} Following Gao \emph{et al.}'s evaluation protocol \cite{gao2021open}, we evaluate COCO-trained models on three unseen datasets: VOC 2007 \cite{everingham2015pascal}, Object365 \cite{Objects365} and LVIS \cite{gupta2019lvis}. To do so, we generate PLs for the novel label spaces of these datasets on the COCO dataset and train a standard Faster R-CNN model. The results of our approach on the three unseen datasets is compared to \cite{gao2021open} in Table~\ref{tab:GTest_PLs}. VL-PLM significantly outperforms \cite{gao2021open} with similar iterations and smaller batch sizes. Note that \cite{gao2021open} requires additional image captions to generate PLs, while VL-PLM can generate PLs for any given category.

\begin{table}[tb]
\begin{center}
\caption{
Open-vocabulary models trained with base categories from COCO are evaluated on unseen datasets. The evaluation protocol follows~\cite{gao2021open} and reports AP50
} \label{tab:GTest_PLs}
    \begin{tabular}{l c c c c}
    \toprule
    {\ PLs} & { \ Iterations$\times$Batch size} & { \ VOC 2007\ } & { \ Object365\ } &{ \  LVIS \ } \\
    \hline
    {Gao \emph{et al.} \cite{gao2021open}} & 150K$\times$64 & 59.2 & 6.9 & 8.0 \\
    {VL-PLM} & 180K$\times$16 & {\bf 67.4} & {\bf 10.9} & {\bf 22.2} \\
    \bottomrule
    \end{tabular}
\end{center}
\end{table}


\subsection{Semi-supervised object detection}
\label{sec:exp_ssod}

In this task, we have annotations for all categories on a small portion of a large image set. This portion is regarded as the labeled set $\mathcal{I}_L$ and the remaining images are regarded as the unlabeled set $\mathcal{I}_U$ i.e. $\mathcal{I}_L \cap \mathcal{I}_U = \varnothing$.

\vspace{1mm}
\noindent \textbf{Experimental setup:} 
Following previous studies \cite{sohn2020detection,xu_iccv2021_softteacher,zhou_cvpr_21}, we conduct experiments on COCO~\cite{COCO} with 1, 2, 5, and 10\% of the training images selected as the labeled data and the rest as the unlabeled data, respectively.
In the supplement, we provide more results for varying numbers of unlabeled data.
To demonstrate how VL-PLM improves PLs for SSOD, we mainly compare our method with the following baselines.
(1) \emph{Supervised}: A vanilla teacher model trained on the labeled set $\mathcal{I}_L$.
(2) \emph{Supervised}+PLs: We apply the vanilla teacher model on the unlabeled set $\mathcal{I}_U$ to generate PLs and train a student model with both ground truth and PLs.
To compare with \emph{Supervised}+PLs, VL-PLM generates PLs for all categories on $\mathcal{I}_U$. Then, those PLs are merged into the PLs from the vanilla teacher as the final PLs to train a student model named as \emph{Supervised}+VL-PLM.
(3) STAC~\cite{sohn2020detection}: A popular SSOD baseline.
To compare with STAC, we only replace its PLs with ours that are used to train \emph{Supervised}+VL-PLM. The new STAC student model is denoted as STAC+VL-PLM.
Here we report the standard metric for COCO, mAP, which is an average over IoU thresholds from 0.5 to 0.95 with a step size of 0.05.

\vspace{1mm}
\noindent \textbf{Implementation details:} 
We follow the same PL generation pipeline and hyperparameters as the OVD experiment, except that we take a class-agnostic Faster R-CNN~\cite{ren_neurips15_fasterrcnn} as our proposal generator and train it on the different COCO splits.
\emph{Supervised} and \emph{Supervised}+PLs are implemented in Detectron2~\cite{wu2019detectron2} and trained for 90,000 iterations with a batch size of 16.
For models related to STAC~\cite{sohn2020detection}, we use the official code of STAC with default settings. 

\vspace{1mm}
\noindent \textbf{Results:} As shown in Table~\ref{table:ssl_results}, models with VL-PLM outperform \emph{Supervised} + PLs and STAC by a clear margin, respectively.
Since the only change to the baselines is the addition of VL-PLM's PLs, we can conclude that V\&L adds clear value to the PLs and can benefit SSOD.
Another interesting finding is that models with VL-PLM provide bigger gains for smaller labeled data, which is the most important regime for SSOD as it brings down annotation costs.
In that regime, PLs from V\&L models are likely stronger than PLs from the small amount of annotated data.
We also want to mention two recent SSOD methods~\cite{xu_iccv2021_softteacher,zhou_cvpr_21} that achieve higher absolute performance, however, only with additional and orthogonal contributions.  VL-PLM may also improve these methods, but here we focus on a fair comparison to other PL-based methods. Moreover, we believe that with better V\&L models, VL-PLM can further improve SSOD.

\begin{table}[tb]
\begin{center}
\caption{Evaluation of pseudo labels for semi-supervised object detection on COCO~\cite{COCO}. 
}
\label{table:ssl_results}
\begin{tabular}{l c c c c}
\toprule
Methods & 1\% COCO & 2\% COCO & 5\% COCO & 10\% COCO\\
\hline
\emph{Supervised} & 9.25 & 12.70 & 17.71 & 22.10 \\
\emph{Supervised}+PLs & 11.18 & 14.88 & 21.20 & 25.98 \\
\emph{Supervised}+VL-PLM & \bf{15.35} & \bf{18.60} & \bf{23.70} & \bf{27.23} \\
\hline
STAC \cite{sohn2020detection} & 13.97 & 18.25 & 24.38 & 28.64 \\
STAC+VL-PLM & {\bf 17.71} & \bf{21.20} & {\bf 26.21} & {\bf 29.61} \\
\bottomrule
\end{tabular}
\end{center}
\end{table}

\subsection{Analysis of pseudo label generation}
\label{subsec:abstudy}
We base our ablation studies on the COCO-ZS setting for OVD unless otherwise specified. All models are trained for 90,000 iterations with a batch size of 16.

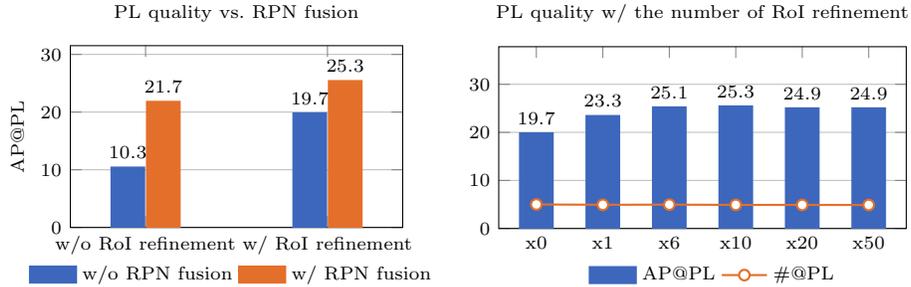
\begin{figure}[t]\centering
\definecolor{fig5orange}{RGB}{228,111,42}
\definecolor{fig5blue}{RGB}{60,103,188}
\subfloat{
\begin{tikzpicture}
\begin{axis}[ybar=.05cm,
    height=4cm,
    width=6cm,
    ymajorgrids,
	bar width=12pt,
	ymin=0,
	ymax=25,
	xtick={1, 2},
	xticklabels={w/o RoI refinement, w/ RoI refinement},
    enlarge x limits={abs=1.0cm},
	enlarge y limits={abs=.5cm, upper},
	ylabel=AP@PL,
	xtick align=inside,
	legend columns=2,
    legend style={at={(-0.15,-0.15)},anchor=north west, legend cell align=left,draw=none},
    title={PL quality vs. RPN fusion},
    nodes near coords,
    point meta=y 
]
\addplot[ybar, draw=fig5blue,
	line width=.4mm,
	fill=fig5blue, area legend
] coordinates {
  (1,10.3) (2,19.7)
};
\addplot[ybar, draw=fig5orange,
	line width=.4mm,
    fill=fig5orange, area legend
] coordinates {
  (1,21.7) (2,25.3)
};
\legend{w/o RPN fusion, w/ RPN fusion}
\end{axis}
\end{tikzpicture}
}%
\subfloat{
\begin{tikzpicture}
\begin{axis}[
    height=4cm,
    width=7cm,
    ymajorgrids,
	bar width=12pt,
	ymin=0,
	ymax=30,
	xtick={1, 2, 3, 4, 5, 6},
	xticklabels={x0, x1, x6, x10, x20, x50},
    enlarge x limits={abs=.5cm},
	enlarge y limits={abs=.5cm, upper},
	xtick align=inside,
	legend columns=2,
    legend style={at={(0.20,-0.15)},anchor=north west, legend cell align=left,draw=none},
    title={PL quality w/ the number of RoI refinement},
]
\addplot[ybar, draw=fig5blue,
	line width=.4mm,
	fill=fig5blue, area legend,
	nodes near coords,
    point meta=y 
] coordinates {
  (1,19.7) (2,23.3) (3,25.1) (4,25.3) (5,24.9) (6,24.9)
};
\addplot[color=fig5orange,solid,thick,mark=*, mark options={fill=white,solid}]
coordinates {
  (1,5.0) (2,4.93) (3,4.96) (4,4.91) (5,4.9) (6,4.89)
};
\legend{AP@PL, \#@PL}
\end{axis}
\end{tikzpicture}
}
    \caption{The quality of PLs with different combinations of RPN and RoI head. We change the threshold $\tau$ to ensure each combination with a similar \#@PL. ``$\times$N'' means we apply RoI head N times to refine the proposal boxes.
    } \label{fig:RPN_RoI_vs_PL_quality}
\end{figure}

\begin{table}[tb]
\begin{center}
\caption{Relationship between the quality of pseudo labels and the performance of the final open vocabulary detectors.}
\label{tab:PL_quality_vs_detector}
\begin{tabular}{l|c|cc|c c c}
\toprule
      & \multirow{2}{*}{PL Setting} & \multicolumn{2}{|c}{Pseudo Labels} & \multicolumn{3}{|c}{Final Detector} \\
\cline{3-7}
      &  & \ AP@PL\  & \ \#@PL\  & \ Base AP\  &\  Novel AP\  & \ Overall AP \\
\hline
\emph{PL v1} & No RoI, $\tau=0.05$ & 17.4 & 89.92 & 33.3 & 14.6 & 28.4 \\
\emph{PL v2} & No RoI, $\tau=0.95$ & 14.6 & 2.88 & 56.1 & 26.0 & 48.2 \\
\hline
\emph{PL v3} & VL-PLM, $\tau=0.05$ & 20.6 & 85.15 & 29.7 & 19.3 & 27.0 \\
\emph{PL v4} & VL-PLM, $\tau=0.95$ & 18.0 & {2.93} & 55.4 & \bf{31.3} & \bf{49.1} \\
\emph{PL v5} & VL-PLM, $\tau=0.99$ & 11.1 & 1.62 & \bf{56.7} & 27.2 & 49.0 \\
\bottomrule
\end{tabular}
\end{center}
\end{table}


\vspace{1mm}
\noindent \textbf{Understanding the quality of PLs: } 
Average precision (AP) is a dominant metric to evaluate object detection methods. 
However, AP alone does not fully indicate the quality of PLs, and the number of PLs also needs to be considered.
To support this claim, we generate 5 sets of PLs as follows. 
(1) \emph{PL v1}: We take the raw region proposals from RPN without RoI refinement in our pseudo label generation and set $\tau = 0.05$. 
(2) \emph{PL v2}: The same as \emph{PL v1} but with $\tau = 0.95$. 
(3) \emph{PL v3}: VL-PLM with $\tau = 0.05$. 
(4) \emph{PL v4}: VL-PLM with $\tau = 0.95$. 
(5) \emph{PL v5}: VL-PLM with $\tau = 0.99$. 
In Table~\ref{tab:PL_quality_vs_detector}, we report AP$_{50}$ (AP@PL) and the average per-image number (\#@PL) of pseudo labels on novel categories. We also report the performance of detection models trained with the corresponding PLs as Novel AP, Base AP and Overall AP.
Comparing \emph{PL v1} with \emph{PL v4} and \emph{PL v2} with \emph{PL v4}, we can see that a good balance between AP@PL and \#@PL is desired. Many PLs may achieve high AP@PL, but drop the performance of the final detector. A high threshold reduces the number of PLs but degrades AP@PL as well as the final performance. We found $\tau = 0.8$ to provide a good trade-off. The table also demonstrates the benefit of VL-PLM over no RoI refinement. 
The supplement contains more analysis and visualizations of our pseudo labels.



\vspace{1mm}
\noindent \textbf{Two-stage proposal generator matters: }
As mentioned in Sec.~\ref{sec:method_pl_with_vl}, we improve the localization ability of CLIP with the two-stage proposal generator in two ways:
1) we merge CLIP scores with RPN scores,
and 2) we repeatedly refine the region proposals from RPN with the RoI Head.
To showcase how RPN and the RoI head help PLs, we evaluate the quality of PLs from different settings in Fig.~\ref{fig:RPN_RoI_vs_PL_quality}.
As shown, RPN score fusion always improves the quality of PLs. As we increase the number of refinement steps with RoI head, the quality increases and converges after about 10 steps. 
Besides proposals from our RPN with RoI refinement (RPN+RoI), we investigate region proposals from different sources, i.e. 1) Selective search~\cite{Uijlings13_selectivesearch}, 2) RPN only, and 3) RoI head with default thresholding and NMS.
Table~\ref{tab:proposals_vs_PL_quality} shows that selective search with a high $\tau$ still leads to a large \#@PL with a low AP@PL for at least two reasons. First, unlike RPN, selective search does not provide objectiveness scores to improve the localization of CLIP. Second, it returns ten times more proposals than RPN, which contain too many noisy boxes.
Finally, the RoI head alone also leads to a poor quality of PLs because it classifies many novel objects as background, due to its training protocol.
In the supplement, we show that the proposal generator, which is trained on base categories, generalizes to novel categories.

\begin{table}[tb]
\begin{center}
\caption{The quality of pseudo labels generated from different region proposals. The threshold $\tau$ is tuned to ensure a similar \#@PL for each method.}
\label{tab:proposals_vs_PL_quality}
\begin{tabular}{l c c c c}
\toprule
     &  Selective Search \cite{Uijlings13_selectivesearch} & \ RoI Head\  & \ RPN\  & \ RPN+RoI (Ours)\  \\
\hline
$\tau$ & 0.99 & 0.55 & 0.88 & 0.82 \\
AP@PL     & 5.7 & 8.8 & 19.7 & \bf{25.3} \\
\#@PL  & 34.92 & 5.01 & 4.70 & \bf{4.26} \\
\bottomrule
\end{tabular}
\end{center}
\end{table}

\vspace{1mm}
\noindent \textbf{Time efficiency: }
VL-PLM sequentially generates PLs for each region proposal, which is time-consuming. For example, VL-PLM with ResNet50 takes 0.54s per image on average. We provide two solutions to reduce the time cost. 
1) Simple multithreading on 8 GPUs can generate PLs for the whole COCO training set within 6 hours.
2) We provide a faster version (Fast VL-PLM) by sharing the ResNet50 feature extraction for all region proposals of the same image. 
This reduces the runtime by $5\times$ with a slight performance drop. Adding multi-scale features (Multiscale Fast VL-PLM) avoids the performance drop but still reduces runtime by $3\times$. Please refer to the supplement for more details.

\section{Conclusion}

This paper demonstrates how to leverage pre-trained V\&L models to mine unlabeled data for different object detection tasks, e.g., OVD and SSOD.  We propose a V\&L model guided pseudo label mining framework (VL-PLM) that is simple but effective, and is able to generate pseudo labels (PLs) for a task-specific labelspace. Our experiments showcase that training a standard detector with our PLs sets a new state-of-the-art for OVD on COCO. Moreover, our PLs can benefit SSOD models, especially when the amount of ground truth labels is limited. We believe that VL-PLM can be further improved with better V\&L models.

\vspace{1mm}
\noindent {\bf Acknowledgments.} 
This research has been partially funded by research grants to D. Metaxas from
NEC Labs America through NSF IUCRC CARTA-1747778, NSF: 1951890, 2003874, 1703883, 1763523 and ARO MURI SCAN.

%
%
\bibliographystyle{splncs04}
\bibliography{egbib.bib}

\newpage
\begin{center}
\textbf{\Large Supplemental Materials}
\end{center}

\appendix

\noindent
The supplemental material first provides additional experiments on open-vocabulary detection (OVD) on the LVIS dataset in Sec.~\ref{subsec:eval_lvis}, a faster version of VL-PLM to speed-up pseudo label (PL) extraction in Sec.~\ref{subsec:fast_vl_plm}, and additional experiments on semi-supervised detection (SSOD) in Sec.~\ref{subsec:more_unlabeled}.
Then, we give additional analysis on fusing pseudo labels in SSOD in Sec.~\ref{subsec:fusion_PL}, the quality of PLs in Sec.~\ref{subsec:quality_PL}, how to model the background category in PL generation in Sec.~\ref{subsec:bg_texts}, and the generalization ability of the proposal generator in Sec.~\ref{subsec:rpn_generalization}.
Finally, qualitative results of PLs and the final OVD detector are given in Sec.~\ref{sec:vis}.

\section{Additional Experiments}

\subsection{Open-vocabulary detection results on LVIS} \label{subsec:eval_lvis}

In addition to our open-vocabulary detection (OVD) experiments on COCO~\cite{COCO} in Sec.~\ref{sec:exp_ovd}, we also evaluate our model on the LVIS~\cite{gupta2019lvis} dataset.
LVIS is a large-vocabulary dataset with 1203 categories and shares images with COCO~\cite{COCO}.
We follow the experimental setup of ViLD~\cite{gu_iclr_22}, the state-of-the-art method on LVIS for OVD (LVIS-OVD):
All categories are divided into three sets, namely, frequent, common, and rare, based on the numbers of their objects.
Following \cite{gu_iclr_22}, we take frequent and common categories as the base categories and regard rare categories as the novel categories.
We leverage base categories to train our two-stage class-agnostic proposal generator and adopt VL-PLM to generate PLs for novel categories. Then, a standard OVD detector was trained with both the ground truth of base categories and our PLs.

\setlength{\tabcolsep}{8pt}
\begin{table}
\begin{center}
\caption{
Evaluations for open vocabulary detection on LVIS-v1~\cite{gupta2019lvis}. 
}
\label{table:open_voc_results_lvis}
    \begin{tabular}{l l cccc}
    \toprule
    Method \ \ & Training data\  & \ AP$_r$\  & \ AP$_c$\  & \ AP$_f$\  & \ mAP\  \\
    \hline
    \emph{Supervised} & Base + Novel & 12.3 & {\bf24.3} & 32.4 & 25.4 \\
    ViLD \cite{gu_iclr_22} & Base & 16.6 & 21.1 &31.6  & 24.4 \\
    VL-PLM (Ours) & Base & {\bf17.2} & 23.7 & {\bf35.1} & {\bf27.0} \\
    \bottomrule
    \end{tabular}
\end{center}
\end{table}
\setlength{\tabcolsep}{1.4pt}

\vspace{1mm}
\noindent \textbf{Comparison with ViLD:} 
Table~\ref{table:open_voc_results_lvis} compares our detector via VL-PLM with \emph{Supervised} and the state-of-the-art method ViLD. \emph{Supervised} is the supervised baseline model trained on the whole LVIS with repeat factor sampling~\cite{gupta2019lvis,mahajan2018eccv_rfs}.
We report the box mAP to better indicate the performance on detection.
For ViLD~\cite{gu_iclr_22}, we took the model provided by the authors and re-ran the evaluation to ensure a fair comparison. For the \emph{Supervised} baseline, we adopted the numbers from \cite{gu_iclr_22}.
As shown, our method outperforms ViLD on all splits. We gain +0.6 AP$_r$ for rare categories (novel) and +2.6 AP$_c$/+3.5 AP$_f$ for common/frequent categories (base). This indicates that training with our PLs has less influence on base categories than the distillation in ViLD does. We observed a similar trend on COCO~\cite{COCO} in Sec.~\ref{sec:exp_ovd}.
Compared with the improvement on base categories, the improvement on novel categories is relatively small, likely due to the long-tailed distribution of novel categories.
Still, VL-PLM outperforms \emph{Supervised} by a large margin in terms of AP$_r$.
A possible explanation is that our PLs provide more annotations for rare categories.
In general, our PLs provide more (but noisy) annotation than the grounded truth ``federated'' annotations of LVIS~\cite{gupta2019lvis}, where only subsets of categories are annotated per image. This may explain why VL-PLM even outperforms \emph{Supervised} in mAP.
Although those annotations are not fully accurate, they still provide useful information for rare categories in the training, e.g., the texture of objects of rare categories.

\subsection{Fast VL-PLM and Multi-scale Fast VL-PLM} \label{subsec:fast_vl_plm}

This section provides more details about Fast VL-PLM and Multi-scale Fast VL-PLM that are mentioned in the discussion of {\bf Time efficiency} in Sec.~\ref{subsec:abstudy}.
Table~\ref{tab:avg_time_PLs} compares original VL-PLM with the two variants in terms of time cost and pseudo label quality. 
As shown, Fast VL-PLM reduces runtime by $5\times$ with a slight drop in PL quality. Multi-scale Fast VL-PLM almost entirely removes the accuracy drop and still reduces runtime by $3\times$.
Our Fast VL-PLM and Multi-scale Fast VL-PLM are only designed for ResNet-based CLIP not for ViT-based CLIP, which are described as follows.

\begin{table}
\begin{center}
\caption{Average time to get pseudo labels per image and their quality.} \label{tab:avg_time_PLs}
    \begin{tabular}{l c c c c}
    \toprule
    {\bf } & { CLIP Backbone} & { \ Time (s)\ } & { \ AP@PL\ } &{ \ \#@PL\ }\\
    \hline
    { Original VL-PLM} & ResNet50 & 0.5413 & 15.9 & 7.31 \\
    { Fast VL-PLM } & ResNet50 & 0.1199 & 13.5 & 7.39 \\
    { Multiscale Fast VL-PLM} & ResNet50 & 0.1685 & 15.4 & 7.38 \\
    \bottomrule
    \end{tabular}
\end{center}
\end{table}

For Fast VL-PLM, we feed the whole input image into CLIP's ResNet50 to get shared feature maps. Then, we bound the coordinates of each region proposal to the close integer. For example, a proposal of $\{10.9, 50.2, 110.1, 100.9\}$ in xyxy format is converted into a box of $\{11, 50, 110, 101\}$. Third, based on the bounded box, we crop features on the shared feature maps for the corresponding proposal.
Finally, we ignore the positional embedding and feed the cropped features into the last attention layer of CLIP to output the region embedding for each proposal. 
Please refer to \cite{radford_arxiv_2021} for details on the structure of ResNet-based CLIP.
Please note the difference to ROI-pooling, where each cropped region would be pooled into the same spatial dimensions. Here, the cropped feature size is proportional to the bounding box and we let the attention layer in CLIP ``pool'' the input into a fixed-size output. We tried ROI-pooling but observed worse performance.

Multi-scale Fast VL-PLM is a mutli-scale version of Fast VL-PLM.
We first construct an image pyramid and feed those images into CLIP's ResNet50 to get shared multi-scale feature maps. 
Then, for region proposals of small size, we crop features on shared feature maps of a large scale so that more details are attained in the cropped feature maps. Shared feature maps of a small scale are for region proposals of large size. 
Specially, we resize the smallest dimension of input images into three scales, i.e., 224, $224 \cdot 3 = 672$, and $224 \cdot 5 = 1120$. Thus, the shared feature maps are in one scale among 7, $7 \cdot 3 = 21$, and $7 \cdot 5 = 35$.
Region proposals are assigned to different scales by their areas. The area $> 64$ is for the first scale, the area between 16 and 64 for the second, and others for the third.
Our design is inspired by FPN \cite{lin_cvpr_17_fpn} and enjoys its advantages, as well.


\subsection{Scaling up unlabeled images for SSOD}
\label{subsec:more_unlabeled}
To better understand the impact of the ratio between labeled and unlabeled images, we continue our experiments on semi-supervised object detection (SSOD). In Sec.~\ref{sec:exp_ssod}, we varied the fraction of labeled images. Here, we use a fixed amount of labeled images and vary the number of unlabeled images.
We train Faster R-CNN models using 5\% of labeled COCO images with different amounts of unlabeled images. 
We randomly select our unlabeled images from the unlabeled images provided by the COCO dataset~\cite{COCO}.
All models are trained for 90k iterations with a batch size of 16. As shown in Table~\ref{tab:AP_data_ratio}, as the amount of unlabeled data increases, the performance increases as well, but with diminishing returns. 
In future work, we want to explore this aspect more and evaluate PLs from VL-PLM in an omni-supervised setting~\cite{radosavovic2018cvpr_omnisupervised}.

\setlength{\tabcolsep}{8pt}
\begin{table}
\begin{center}
\caption{
Detection accuracy in mAP for Faster R-CNN on 5\% of labeled COCO images with varied amounts of unlabeled images.
}
\label{tab:AP_data_ratio}
    \begin{tabular}{l |c|c|c|c|c}
    \toprule
    \# of labeled images & 5764 & 5764 & 5764 & 5764 & 5764 \\
    \# of unlabeled images & 0 & 5764 & 28820 & 57640 & 115280 \\
    Ratio      & 1:0 & 1:1 & 1:5 & 1:10 & 1:20 \\
    \hline
    mAP & 17.7 & 21.1 & 22.7 & 23.1 & 23.7 \\
    mAP increase & - & +3.4 & +5.0 & +5.4 & +6.0 \\
    \bottomrule
    \end{tabular}
\end{center}
\end{table}
\setlength{\tabcolsep}{1.4pt}

\section{Additional Analysis and Discussion}

\begin{figure}[b]\centering
    \includegraphics[width=.7\linewidth]{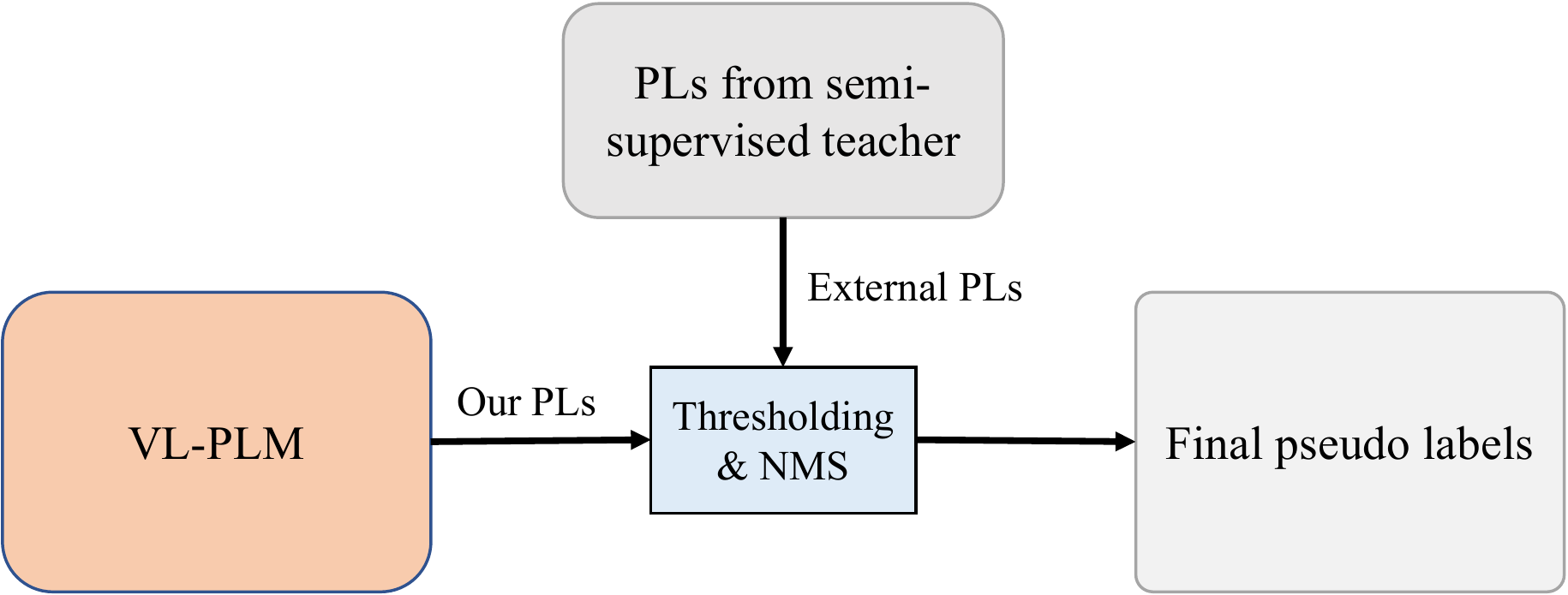}
    \caption{Overview of pseudo labels (PLs) fusion for semi-supervised object detection (SSOD). We fuse our PLs with those from the semi-supervised teacher model before the thresholding and NMS. }
    \label{fig:ssod_pls}
\end{figure}

\subsection{Fusing Pseudo Labels from SSOD teacher with VL-PLM}\label{subsec:fusion_PL}

This section provides more details for Sec.~\ref{sec:method_pl_for_specific_tasks} of the main paper on how we merge PLs for SSOD.
As illustrated in Fig.~\ref{fig:ssod_pls}, we use the proposed VL-PLM to generate PLs and merge the PLs from the semi-supervised teacher. Then, we apply thresholding and NMS on the merged PLs to obtain the final PLs for SSOD.
To validate the effectiveness of this fusion strategy, we consider the following baselines. 
(1) SSL PLs only: We only adopt the PLs from the semi-supervised teacher as the final PLs.
(2) VL-PLM w/o fusion: We only pick the PLs from the vision and language model.
(3) VL-PLM: The fused PLs.
We base our experiments on 1\%, 2\%, 5\% and 10\% COCO splits \cite{COCO,sohn2020detection} for SSOD and report the results in Table~\ref{table:ssod_pl_fusion}. 
As shown, compared with SSL PLs only, VL-PLM w/o fusion is better on 1\% and 2\% COCO splits but worse on 5\% and 10\% COCO splits. A possible explanation is that V\&L models provide more useful information that boosts the performance when the amount of annotated data is smaller.
Moreover, VL-PLM outperforms SSL PLs only and VL-PLM w/o fusion on all splits. This clearly demonstrates that our PLs from VL-PLM are better than the PLs from the teacher model. Our fusion method successfully improves the quality of the final PLs.
Since putting PLs from the teacher and V\&L model together brings about the best results even for 5\% and 10\% COCO splits, we believe that PLs from the teacher and the V\&L model are complementary.

\begin{table}[tb]
\begin{center}
\caption{Pseudo label fusion for semi-supervised object detection on COCO 2017 \cite{COCO}. 
}
\label{table:ssod_pl_fusion}
\begin{tabular}{l c c c c}
\toprule
Methods\  & \ 1\% COCO\  & \ 2\% COCO\  & \ 5\% COCO\  & \ 10\% COCO\ \\
\hline
SSL PLs only       & 11.18 & 14.88 & 21.20 & 25.98 \\
VL-PLM w/o fusion  & 13.27 & 15.97 & 20.64 & 24.20 \\
VL-PLM             & \bf{15.35} & \bf{18.60} & \bf{23.70} & \bf{27.23} \\
\bottomrule
\end{tabular}
\end{center}
\end{table}

\subsection{Analysis on the Quality of PLs} \label{subsec:quality_PL}
In this section, we provide a more detailed analysis and discussion for {\bf Understanding the quality of PLs} of Sec.~\ref{subsec:abstudy} in the main paper.

\begin{table}[tb]
\begin{center}
\caption{Relationship between the quality of pseudo labels and the performance of the final open vocabulary detectors on COCO 2017 \cite{COCO}.}
\label{tab:sup_PL_quality_vs_detector}
\begin{tabular}{l|c|cc|c c c}
\toprule
      & \multirow{2}{*}{PL Setting} & \multicolumn{2}{|c}{Pseudo Labels} & \multicolumn{3}{|c}{Final Detector} \\
\cline{3-7}
      &  & \ AP@PL\  & \ \#@PL\  & \ Base AP\  &\  Novel AP\  & \ Overall AP \\
\hline
\emph{PL v1} & No RoI, $\tau=0.05$ & 17.4 & 89.92 & 33.3 & 14.6 & 28.4 \\
\emph{PL v2} & No RoI, $\tau=0.95$ & 14.6 & 2.88 & 56.1 & 26.0 & 48.2 \\
\hline
\emph{PL v3} & VL-PLM, $\tau=0.05$ & 20.6 & 85.15 & 29.7 & 19.3 & 27.0 \\
\emph{PL v4} & VL-PLM, $\tau=0.95$ & 18.0 & {2.93} & 55.4 & \bf{31.3} & \bf{49.1} \\
\emph{PL v5} & VL-PLM, $\tau=0.99$ & 11.1 & 1.62 & \bf{56.7} & 27.2 & 49.0 \\
\bottomrule
\end{tabular}
\end{center}
\end{table}

\vspace{1mm}
\noindent \textbf{Quality of PLs and performance of final detectors:} 
In this section, we provide more analysis for Table~\ref{tab:sup_PL_quality_vs_detector} which is also present in the main paper. We recall our 5 baselines as follows, 
(1) \emph{PL v1}: We take the raw region proposals from region proposal network (RPN) without RoI refinement in our pseudo label generation and set $\tau = 0.05$. 
(2) \emph{PL v2}: The same as \emph{PL v1} but with $\tau = 0.95$. 
(3) \emph{PL v3}: VL-PLM with $\tau = 0.05$. 
(4) \emph{PL v4}: VL-PLM with $\tau = 0.95$. 
(5) \emph{PL v5}: VL-PLM with $\tau = 0.99$. 
In Table~\ref{tab:sup_PL_quality_vs_detector}, the evaluations are conducted on the zero-shot splits \cite{bansal2018zero} on COCO \cite{COCO} (COCO-ZS). We report the AP$_{50}$ (AP@PL) and the number (\#@PL) on novel categories for different PLs with the performance of detection models trained with corresponding PLs. Novel AP, Base AP, and Overall AP are provided to indicate the performance of detectors. 

Based on Table~\ref{tab:sup_PL_quality_vs_detector}, we have the following findings. 
First, compared with \emph{PL v4}, \emph{PL v1} shares similar AP@PL but has much more pseudo annotations. The final detector of \emph{PL v4} significantly outperforms that of \emph{PL v1}.
Second, compared with \emph{PL v4}, \emph{PL v2} has nearly the same amount of PLs with a lower AP@PL. In terms of Novel AP, the final detector of \emph{PL v4} outperforms that of \emph{PL v1} by a large margin.
Based on those facts, we conclude that neither AP@PL nor \#@PL alone can decide the quality of PLs. We need to consider both AP@PL and \#@PL. Good PLs come with high AP@PL and low \#@PL. 
Third, based on the comparison between \emph{PL v4} and \emph{PL v5}, we find that an extremely high threshold $\tau$ harms the predictions of novel categories but results in a slightly better performance on base categories. Empirically, we find a reasonable $\tau \in [0.6,0.95]$ and set $\tau = 0.8$ as default.
Fourth, comparing \emph{PL v2} and \emph{PL v4}, we find that with RoI head refinement, our PLs gain a significant improvement. For the final detector, \emph{PL v4} achieves similar performance on base categories as \emph{PL v2} and much better results on novel categories, which clearly demonstrate the effectiveness of using RoI head as box refinement.

\begin{table}[tb]
\begin{center}
\caption{The quality of pseudo labels generated with different ways to model the background. $\tau$ is tuned to keep similar \#@PL. See text for more details.}
\label{tab:PL_quality_vs_bg}
\begin{tabular}{l c c c c c}
\toprule
      & Novel\  & \ Novel+BG\  & \ Novel+Base\  & \ Novel+Base+BG\  & \ Novel+OV set\ \\
\hline
$\tau$ & 0.80 & 0.80 & 0.59 & 0.59 & 0.501 \\
AP@PL & 25.5 & 25.7 & 26.3 & 26.4 & 25.1 \\
\#@PL & 4.86 & 4.18 & 4.36 & 4.22 & 4.35 \\
\bottomrule
\end{tabular}
\end{center}
\end{table}

\subsection{Modeling Background in PL Generation} \label{subsec:bg_texts}
Background is a latent category for the detection task and should be considered in our pseudo label generation, as well. In this section, we demonstrate how different ways of modeling the background affects the quality of PLs for OVD on COCO-ZS. Since there may be region proposals for base categories, we generalize the concept of background as categories that are not in the target categories, and consider 5 category spaces with different backgrounds as
\begin{enumerate}
    \item \emph{Novel}: The label space for pseudo labels only contains novel categories, no explicit modeling of background
    \item \emph{Novel+BG} The text ``background'' is used as one additional background category
    \item \emph{Novel+Base}: Both novel and base categories are used in the label space of PLs, where base categories should model the background (since those are annotated in the OVD setting)
    \item \emph{Novel+Base+BG}: Same as \emph{Novel+Base}, but with the additional category of ``background''
    \item \emph{Novel+OV}: We remove novel categories in COCO from the 1203 categories in LVIS~\cite{gupta2019lvis}. The remaining categories are used to model background. We name this background set as OV.
\end{enumerate}
Table~\ref{tab:PL_quality_vs_bg} provides AP@PL and \#@PL on novel categories for different category spaces of the background.

As shown, \emph{Novel+BG} is slightly better than \emph{Novel} with higher AP@PL and lower \#@PL. \emph{Novel+Base} and \emph{Novel+Base+BG} result in the same observation. BG does improve the quality but the improvement is not significant. Moreover, \emph{Novel+Base} gains a clear improvement over \emph{Novel}, likely because it helps V\&L models to identify objects of base categories which will be removed as the background, improving the quality of PLs for novel categories. Third, OV as the background decreases the quality of PLs based on the comparison between\emph{Novel+OV} and \emph{Novel+Base}. This is reasonable because V\&L models may be influenced by the large amount of OV categories that are absent in the scene.


\subsection{Generalization ability of the proposal generator} \label{subsec:rpn_generalization}
For OVD, we need to identify the objects of novel categories in the unlabeled data.
Similar to \cite{gu_iclr_22}, we study if the two-stage class-agnostic detector trained on base categories generalizes to novel ones. 
Basing our experiments on COCO and LVIS datasets, we train a Faster R-CNN with either base or all categories (base+novel). 
Table~\ref{tab:sup_proposal_ar} presents the top-N average recall (AR@N) of the RPN on novel categories.
As shown, on COCO-ZS, RPN trained without novel categories suffers a clear performance drop. But it can still achieve a reasonable AR@1000, and we adopt top 1000 boxes from RPN in our pseudo label generation.
On the contrary, models on LVIS achieve nearly the same average recall (AR) on novel categories with either base categories or base + novel categories as the training data. Possibly, there are more categories and instances in base categories of LVIS than in those of COCO. Thus, the proposal generator is able to learn a better concept about the objects in LVIS.

\begin{table}
\begin{center}
\caption{The generalization ability of RPN of the proposal generator on COCO 2017 \cite{COCO} and LVIS-v1 \cite{gupta2019lvis}. We report the average recall (AR) on novel categories.}
\label{tab:sup_proposal_ar}
\begin{tabular}{l l c c c c}
\toprule
Dataset\ \  & Training Data\  & \ AR@100\  & \ AR@300\  & \ AR@500\  & \ AR@1000\  \\
\hline
\multirow{2}{*}{COCO} & Base & 34.5 & 43.4 & 47.2 & 51.7 \\
                      & Base + Novel  & 54.2 & 59.3 & 61.1 & 62.8 \\
\hline
\multirow{2}{*}{LVIS} & Base & 33.3 & 42.3 & 45.9 & 50.5 \\
                      & Base + Novel  & 33.7 & 43.0 & 46.6 & 50.5 \\
\bottomrule
\end{tabular}
\end{center}
\end{table}

\begin{figure}[t]\centering
    \includegraphics[width=1\linewidth]{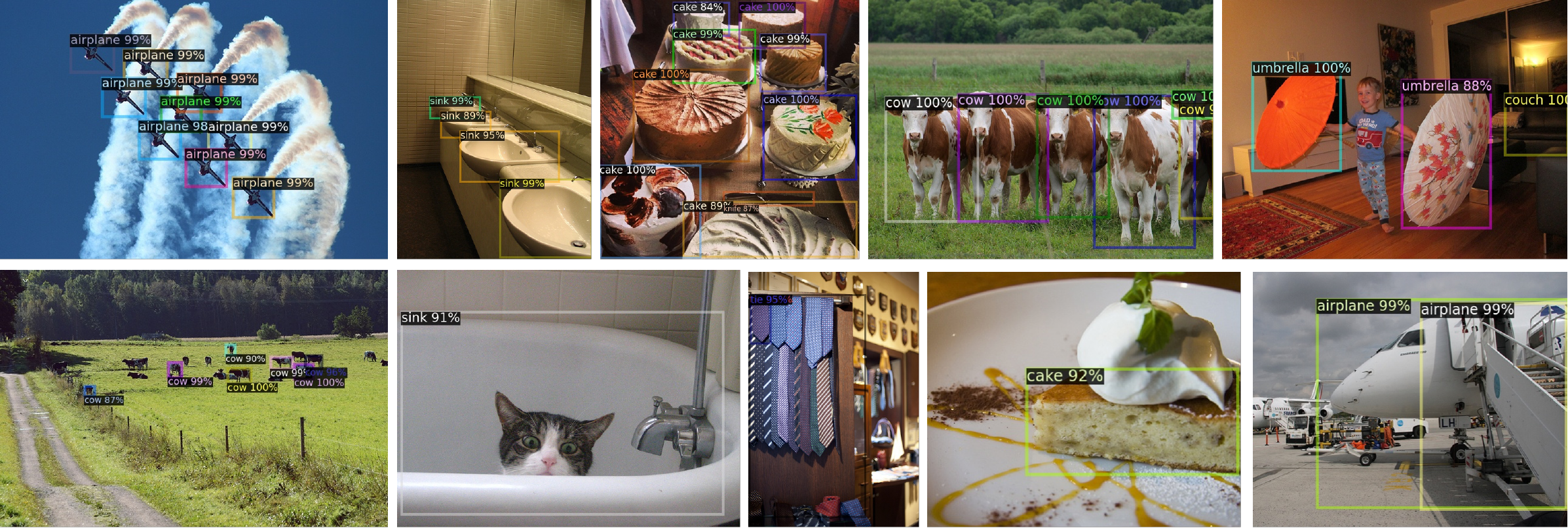}
    \caption{Visualization of pseudo labels. Only novel categories in the image are shown. Top: Good cases with multiple instances. Bottom: Failure cases with missing instances, grouped instances, part domination and redundant annotations.}
    \label{fig:pl_visualization}
\end{figure}

\begin{figure}[t]\centering
    \includegraphics[width=.96\linewidth]{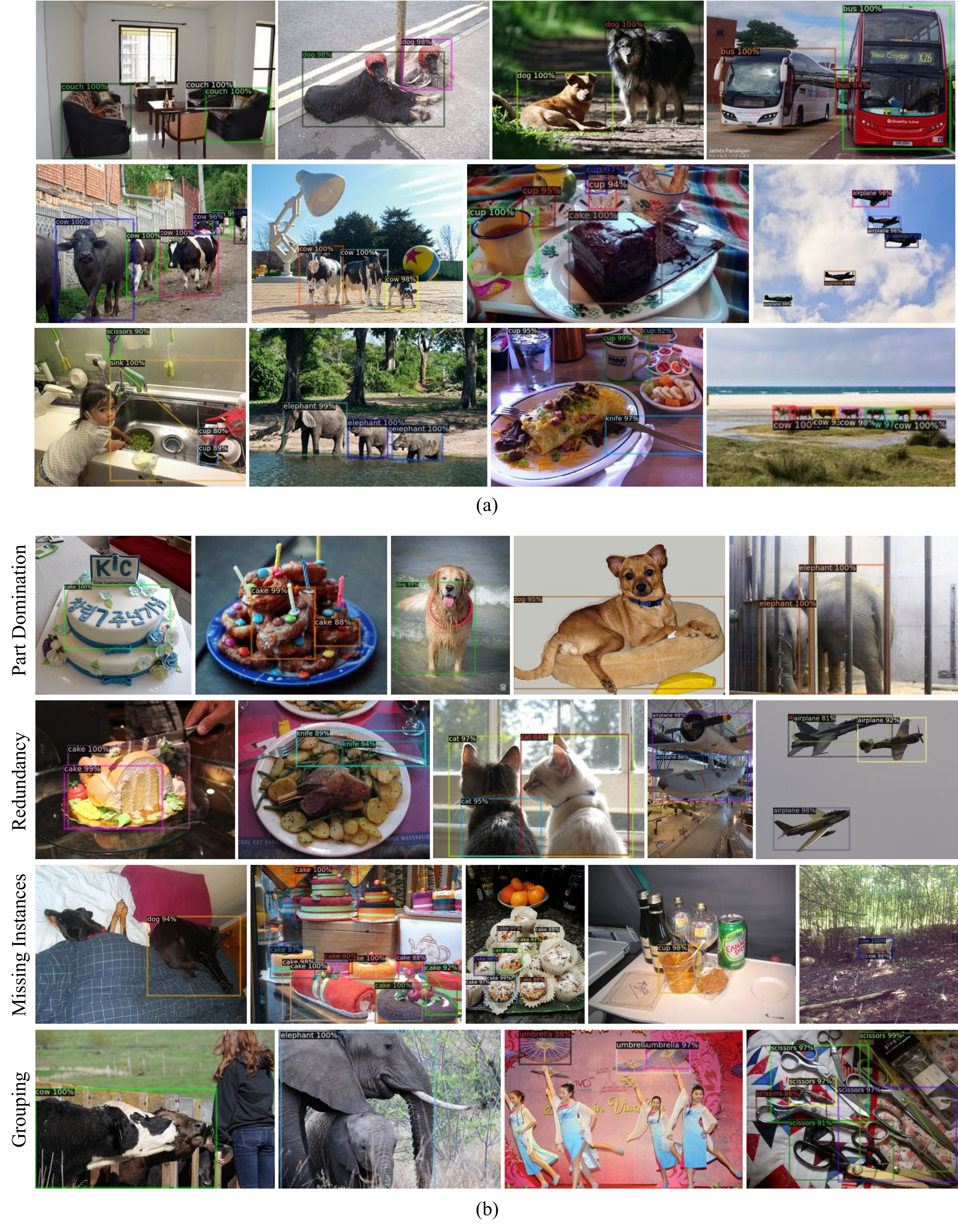}
    \caption{Visualizations of the pseudo labels (PLs) from VL-PLM. Only boxes for target categories in the scene are shown. (a) Good cases. All target objects are located with appropriate boxes. (b) The most common types of failure cases in our PLs, i.e., part domination, redundant boxes, missing instances, and grouped instances.}
    \label{fig:pl_vis}
\end{figure}

\begin{figure}[t]\centering
    \includegraphics[width=.96\linewidth]{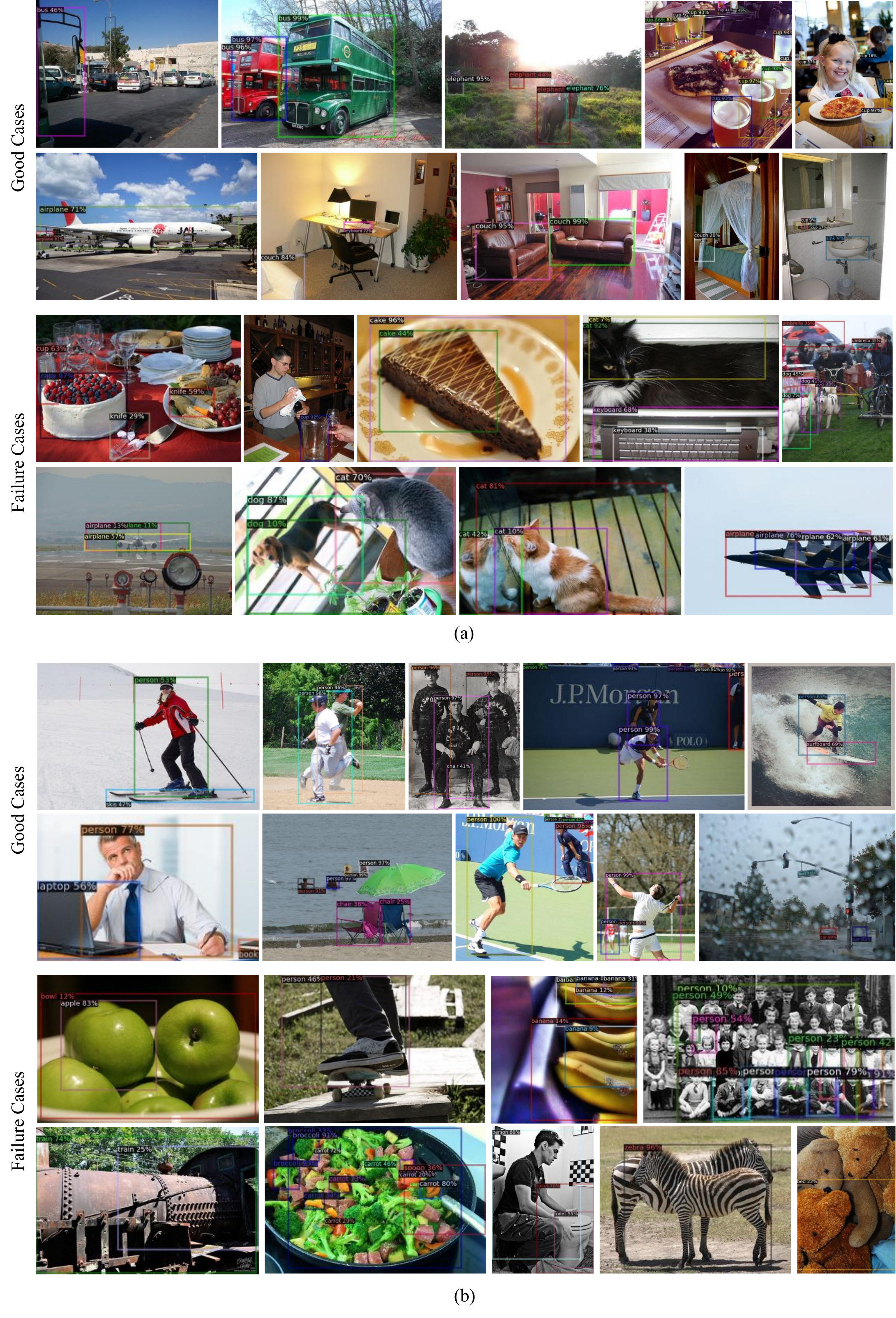}
    \vspace{-0.4cm}
    \caption{Visualization of the final detection results. Only boxes for target categories in the scene are shown. (a) Novel categories as the target. (b) Base categories as the target. The major failure cases belong to three types, i.e., missing instances, redundant boxes, or grouped instances.}
    \label{fig:detector_vis}
\end{figure}

\section{Qualitative Results} \label{sec:vis}

\subsection{Visualization of Our Pseudo Labels}

We illustrate good cases and failure cases of our PLs for OVD on COCO in Fig.~\ref{fig:pl_visualization} and Fig.~\ref{fig:pl_vis}. We only visualize boxes for target (novel) categories that are included in the scene. 
Good cases show that VL-PLM is able to locate multiple objects correctly. However, in the recent caption-based pseudo label generation method \cite{gao2021open}, it's a major issue to find multiple objects of the same categories.
For failure cases, there are four major types, i.e., part domination, redundant boxes, missing instances, and grouped instances. 
We believe that part domination and redundant boxes are mainly caused by the poor localization ability of the adopted V\&L model CLIP \cite{radford_arxiv_2021}. 
Missing and grouped instances usually happen when multiple instances are close to each other or only part of instances appears. Possibly, the major reason is that the proposal generator cannot provide correct region proposals, leading to poor quality of PLs. 
In this sense, an improvement on either the V\&L models or the proposal generator in VL-PLM will boost PLs' quality.

\subsection{Visualization of Our OVD Detector}

This section visualizes the good and failure cases of the final detector for OVD. Fig.~\ref{fig:detector_vis} illustrates those cases on novel and base categories, respectively. 
As shown, the detector trained with our PLs is able to detect objects of novel categories. Moreover, unlike PLs, the results of the final detector mainly include three types of failure cases, i.e., missing instances, redundant boxes, and grouped instances. Possibly, the part domination that degrades the quality of PLs is reduced during the training with both ground truth and PLs.

\end{document}